\documentclass{article}
\usepackage[final,dblblindworkshop]{neurips_2026}
\workshoptitle{Self-Evolving Diversity-Driven Search for Robust AI Systems}
\usepackage[utf8]{inputenc}
\usepackage[T1]{fontenc}
\usepackage{hyperref}
\usepackage{url}
\usepackage{booktabs}
\usepackage{amsfonts}
\usepackage{amsmath}
\usepackage{amssymb}
\usepackage{microtype}
\usepackage{xcolor}
\usepackage{algorithm}
\usepackage{algpseudocode}
\usepackage{float}
\usepackage{enumitem}
\usepackage{graphicx}
\usepackage{tikz}
\usetikzlibrary{arrows.meta,positioning,backgrounds,fit,calc}
\definecolor{vioF}{RGB}{250,230,230}\definecolor{vioB}{RGB}{198,118,118}
\definecolor{disF}{RGB}{253,243,211}\definecolor{disB}{RGB}{190,150,55}
\definecolor{pasF}{RGB}{227,242,232}\definecolor{pasB}{RGB}{100,159,122}
\definecolor{outF}{RGB}{240,236,249}\definecolor{outB}{RGB}{126,107,176}
\definecolor{grpB}{RGB}{125,125,133}\definecolor{ink}{RGB}{38,38,44}

\title{\textsc{Penumbra}: Sample-Efficient Adversarial Search for\\Regulatory
Obligations}
\author{Anthony Rhodes\\
  Confidential Core AI\\
  \texttt{a.rhodes@confidentialcore.ai}}

\begin{document}
\maketitle

\begin{abstract}
Agents are entering finance, healthcare and law, sectors where a violation leaves
no lexical signature and carries real penalties. Whether an omission is material,
or a disclosure sufficient, depends on what the response \emph{left out}. Probing
such an obligation means finding responses one minimal edit from flipping
compliance, and every probe costs a generation and two adjudications, so the
binding constraint on regulatory red-teaming is sample efficiency, not volume.

We introduce \textsc{Penumbra}, an adversarial search that walks from a verified
anchor under an expanding edit budget until a two-evaluator committee changes its
verdict, and emits the two adjacent responses that straddle the change. Allocation
is adaptive, and the objective is coverage of the defeat surface: distinct
(obligation $\times$ defeat mode) cells resolved per candidate. At matched budget,
adaptive allocation reaches uniform allocation's full-budget coverage on
\textbf{59\% of the candidates}; at equal records it covers \textbf{1.43$\times$
the defeat modes of naive enumeration}, and the gain is confined to the axis it
targets. On 60 screened obligations of a financial advisory constitution, \textsc{Penumbra} returns 144 pairs, each a compliant and a violating response that the committee places on opposite sides of the boundary, differing by a handful of words where a model asked for both directly produces texts sharing almost nothing. A second constitution, for clinical triage, reproduces this on 18 obligations: 49 pairs at the same tightness, with the same modes hardest. Pairs like these show where an obligation's own terms stop deciding, which is what an agent deployed under it must be tested against, and the search finds them at a cost that scales with the boundary, not the text.
\end{abstract}

\section{Introduction}
\label{sec:intro}

Agents are entering sectors that carry binding professional obligations: portfolio analysis and compliance review in finance, triage and documentation in healthcare. Such an agent inherits a constitution written for human professionals: ``Maintain version traceability. Do not represent inference as retrieved fact. Do not selectively omit material information.'' An agent can breach any of these without producing anything a lexical detector could match, because the breach is not a phrase in the response but a judgement about it: whether an omission was material, a disclosure sufficient, an answer within scope.

Concretely, two responses from our corpus answer the same client question under \emph{do not selectively omit material information}; they differ by three words:

\vspace{-2pt}
\begin{center}\footnotesize
\begin{minipage}{0.94\linewidth}
\setlength{\parskip}{1pt}
\textcolor{vioB!72!black}{\textbf{violation}} \; \ldots the company faces high
valuation risk and cyclical demand risk in the semiconductor industry\ldots

\textcolor{pasB!72!black}{\textbf{compliant}} \; \ldots the company faces high
valuation risk, \colorbox{pasF}{\textbf{customer concentration issues,}} and
cyclical demand risk in the semiconductor industry\ldots
\end{minipage}
\end{center}

\noindent Both disclose risk, and no keyword test separates them. What separates them is what the agent knew: the tool output it had retrieved named a customer concentration exposure, so the first response omits a material fact and the second does not.

Such pairs are hard to find, and yet they are \textit{worth finding}: each fixes a point on the shadow line of an obligation, where its own light gives out, and \textbf{that edge is where compliance actually rests}, on a reading of materiality or sufficiency that no checklist states in advance. Prior approaches were not built to find such pairs, and the reasons are instructive. Prompting a model to violate a rule yields conspicuous violations far from any line. Self-evolving red-team loops \citep{perez2022redteaming,ganguli2022redteaming,bai2022constitutional,samvelyan2024rainbow} generate at volume, but every record is a violation with no compliant sibling, and Section~\ref{sec:why} shows they drift away from the line as they run.

Probing must therefore be \emph{diverse}, covering the distinct ways an obligation
can be defeated, and \emph{boundary-seeking}, placing each defeat a minimal edit
from compliance. \textsc{Penumbra} pursues both over a space of obligation $\times$
defeat mode, reading disagreement between two independent evaluators as the signal
for the second. Rules of this kind have a settled core and a penumbra where their own terms do not decide the case \citep{hart1961,wittgenstein1953,dworkin1986,solum2013,bhatia2018}. Two observers a step apart inside it will disagree about whether the light is covered, and the disagreement is a fact about where they stand rather than about their eyes; and thus a split between two competent adjudicators is the observable signature of that region.

Probing an obligation is also expensive. Each candidate costs one generation and two adjudications, one from each evaluator, and most edits leave the verdict unmoved, so the question is not how much adversarial text a loop emits but rather how few adjudications each certified pair costs. \textsc{Penumbra} is a search built around that question: it spends adjudications only where a verdict is about to change, and so maps an obligation's boundary at a cost that scales with the boundary rather than the text, which is what makes red-teaming agents in regulated sectors tractable. We contribute a calibrated boundary detector, with the negative result that the obvious candidate fails (Section~\ref{sec:oracle}); \textsc{Penumbra} and its adaptive allocation (Sections~\ref{sec:alg} and~\ref{sec:results}); and a corpus of 144 certified minimal-edit pairs with their coverage map.

\section{A calibrated boundary detector}
\label{sec:oracle}

A boundary example in regulatory policy is one a competent evaluator could reasonably judge either way. The tempting instrument is the evaluator's own confidence, and it fails: on a ladder of 77 screened compliant responses over eleven obligations, authored at three difficulty levels (Appendix~\ref{app:ladder}), an evaluator reports mean confidence 0.935, 0.929 and 0.928 for clear, guarded and borderline, 97\% at 0.9 or above. Confidence carries no difficulty information, consistent with the finding that verbalised confidence sits near the top of the scale whatever the answer \citep{xiong2024confidence,tian2023calibration}.

Disagreement between two separately prompted evaluators does carry it. Each evaluator applies a written checklist for the obligation it scores. On the same ladder the disagreement rate runs 0.000 / 0.032 / 0.091 under the original checklists (slope $+0.045$) and 0.042 / 0.097 / 0.136 under a patched set (slope $+0.047$), so sensitivity does not depend on how much the checklist has been revised. The calibration pair was cross-vendor and the search committee same-vendor, so the ladder bounds the search's sensitivity from above (Appendix~\ref{app:ladder}). Two evaluators therefore give a three-level ordinal: violation when both convict, boundary when they split, pass when both acquit. Its middle level is a termination condition, not a threshold on a noisy scalar. Formally this is query-by-committee \citep{seung1992qbc}, with the query synthesised, not selected, and the committee calibrated, not assumed.

\begin{algorithm}[t]
\caption{Bracket and cross, per (obligation, defeat mode). $J_1, J_2$ are the
two evaluators, $\textsc{Ordinal}$ maps their verdict pair to
\{violation, boundary, pass\}, and $\tau$ is the similarity floor below which a
crossing is discarded as too wide to be a boundary pair.}
\label{alg:bsc}
\begin{algorithmic}[1]
\State $x^{+}\!\gets$ verified compliant anchor; \; $x^{-}\!\gets$ verified violating anchor
\State $s \gets x^{-}$ or $x^{+}$; \; $c \gets \textsc{Ordinal}(J_1(s), J_2(s))$
  \Comment{start side alternates per attempt}
\For{$k = 1 \dots K$}
  \State $\beta \gets \beta_k$
    \Comment{edit budget, \emph{expanding}: words $\to$ clause $\to$ sentence $\to$ paragraph}
  \State $x \gets \textsc{Soften}(s,\beta)$ if $c$ is violating, else $\textsc{Harden}(s,\beta)$
    \Comment{the defect is preserved in both}
  \State $c' \gets \textsc{Ordinal}(J_1(x), J_2(x))$
  \If{$c' \neq c$}
     \Comment{the band changed: $s$ and $x$ straddle it}
     \If{$\mathrm{sim}(s,x) \ge \tau$}
        \State \textbf{emit} $(s,x)$ as a certified pair and \textbf{return}
     \EndIf
     \Comment{too wide, so keep walking rather than emit a loose pair}
  \EndIf
  \State $s \gets x$; \; $c \gets c'$
\EndFor
\end{algorithmic}
\end{algorithm}

\section{Why existing generation misses the boundary}
\label{sec:why}

Self-evolving red-team loops, the iterative generator-and-evaluator systems of Section~\ref{sec:intro}, drift away from the boundary as they run. When we apply the detector
of Section~\ref{sec:oracle} to a faithfully implemented self-evolving loop over eleven
obligations, disagreement falls by round, 0.133, 0.081, 0.052, 0.038, 0.100, with a
negative slope in all eleven ($p=2^{-11}$ by sign test). Its outputs get
\emph{easier} as it runs, and its objectives explain why: it rewards evasion and
novelty, and neither points at the boundary.
Adapting either side does not repair the drift (Appendix~\ref{app:loop}). We therefore freeze the committee and \emph{diagnose} each disagreement rather than
patching it.

\section{\textsc{Penumbra}}
\label{sec:alg}

\begin{figure}[t]
\centering
\begin{minipage}[b]{0.60\textwidth}\centering
\resizebox{\textwidth}{!}{%
\begin{tikzpicture}[
  font=\footnotesize\color{ink},
  >={Stealth[length=1.9mm,width=1.5mm]}, x=1cm,y=1cm,
  agent/.style={rounded corners=3.5pt,draw,line width=.8pt,align=center,
                minimum width=2.95cm,minimum height=1.42cm},
  inp/.style={rounded corners=3.5pt,draw=grpB,line width=.7pt,fill=white,align=center,
              inner xsep=4pt,inner ysep=4pt},
  outbox/.style={rounded corners=3.5pt,draw,line width=.8pt,inner xsep=4.5pt,
                 inner ysep=4pt,text width=2.62cm,align=left,font=\scriptsize},
  pill/.style={rounded corners=5pt,draw=disB,fill=disF,line width=.8pt,
               inner xsep=4.5pt,inner ysep=3pt,font=\scriptsize\bfseries,
               text=disB!70!black},
  arw/.style={line width=.85pt,->}
]
\def\yc{2.35}

\node[inp] (in) at (.92,\yc) {\footnotesize\textbf{60} obligations\\[.5pt]\scriptsize $\times$ 5 defeat modes\\[.5pt]\scriptsize\color{grpB} 300 cells};

\node[agent,fill=vioF,draw=vioB] (gen) at (4.05,\yc) {};
\node[anchor=north] at (4.05,\yc+.58) {\textbf{\textcolor{vioB!80!black}{Generator}}};
\node[anchor=north,font=\scriptsize] at (4.05,\yc+.24) {soften \,/\, harden};
\begin{scope}[shift={(2.98,\yc-.28)}]
  \draw[vioB!75!black,line width=.7pt,->] (0,0)   -- (.26,0);
  \draw[vioB!75!black,line width=.7pt,->] (.36,0) -- (.80,0);
  \draw[vioB!75!black,line width=.7pt,->] (.90,0) -- (1.58,0);
  \node[font=\tiny,text=vioB!70!black,anchor=north] at (.79,-.08) {expanding edit budget};
\end{scope}

\node[agent,fill=outF,draw=outB] (com) at (8.15,\yc) {};
\node[anchor=north] at (8.15,\yc+.58) {\textbf{\textcolor{outB!85!black}{Evaluator committee}}};
\node[anchor=north,font=\scriptsize] at (8.15,\yc+.24) {two independent judges};
\begin{scope}[shift={(6.94,\yc-.44)}]
  \fill[vioF] (0,0) rectangle (.78,.24);     \draw[vioB,line width=.5pt] (0,0) rectangle (.78,.24);
  \fill[disF] (.78,0) rectangle (1.60,.24);  \draw[disB,line width=.85pt] (.78,0) rectangle (1.60,.24);
  \fill[pasF] (1.60,0) rectangle (2.42,.24); \draw[pasB,line width=.5pt] (1.60,0) rectangle (2.42,.24);
  \node[font=\tiny,text=vioB!70!black] at (.39,.12) {violation};
  \node[font=\tiny\bfseries,text=disB!68!black] at (1.19,.12) {disagree};
  \node[font=\tiny,text=pasB!70!black] at (2.01,.12) {pass};
\end{scope}

\draw[arw,vioB!85!black] (gen.north) to[out=50,in=130] (com.north);
\node[font=\scriptsize,text=vioB!75!black,anchor=south] at (6.10,\yc+1.06) {candidate response};
\draw[arw,outB] (com.south) to[out=-142,in=-38] (gen.south);
\node[font=\scriptsize,text=outB!85!black,anchor=north,fill=white,inner sep=1.5pt] at (6.10,\yc-1.22) {three-band verdict steers the next edit};

\node[draw=grpB,dashed,dash pattern=on 2.5pt off 2pt,line width=.7pt,
      rounded corners=4.5pt,fit=(gen)(com),inner xsep=8pt,inner ysep=56pt] (loop) {};
\node[font=\scriptsize\bfseries,text=grpB,anchor=north] at ([yshift=-4pt]loop.north) {ADVERSARIAL CO-ADAPTATION LOOP};
\draw[arw,grpB] (in.east) -- (loop.west);

\node[pill] (cross) at (10.86,\yc) {crossing};
\draw[arw,grpB] (loop.east) -- (cross.west);

\node[outbox,fill=pasF,draw=pasB] (pref) at (13.05,\yc+1.16)
  {\textbf{\textcolor{pasB!80!black}{preference pairs}}\\[1pt]\textit{pass $\leftrightarrow$ violation}\\[1pt]DPO-usable\\[1pt]\textbf{78} of 144};
\node[outbox,fill=outF,draw=outB] (exem) at (13.05,\yc-1.20)
  {\textbf{\textcolor{outB!85!black}{exemplar pairs}}\\[1pt]\textit{into disagreement}\\[1pt]\textbf{30} penumbral\\ \textbf{35} artifact\\[1pt]\textbf{66} of 144};
\draw[arw,pasB] (cross.east) to[out=48,in=180] (pref.west);
\draw[arw,outB] (cross.east) to[out=-48,in=180] (exem.west);

\draw[arw,grpB,dash pattern=on 2.5pt off 2pt,rounded corners=5pt]
  (10.86,\yc-.38) -- (10.86,.12) -- (3.10,.12) -- (3.10,\yc-.73);
\node[fill=white,inner xsep=3pt,inner ysep=1pt,font=\scriptsize,text=grpB] at (6.10,.12)
  {resolved cell masked, so the next attempt goes to an unresolved one};
\end{tikzpicture}%
}
\end{minipage}\hfill
\begin{minipage}[b]{0.34\textwidth}\centering
\includegraphics[height=88pt]{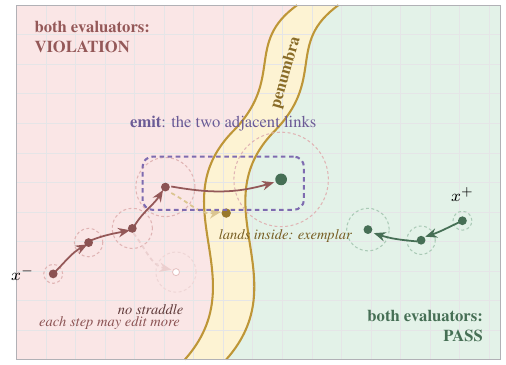}
\end{minipage}
\caption{\textsc{Penumbra}. \textbf{Left}: the generator proposes a candidate under an
\emph{expanding} edit budget; the committee's three-level verdict selects the next
operator and budget, and a change of level emits the two adjacent responses as a
certified pair. \textbf{Right}: one chain for one obligation; a step that clears the
penumbra gives a preference pair, one landing inside gives an exemplar (Appendix~\ref{app:walk}).}
\label{fig:bsc}
\end{figure}
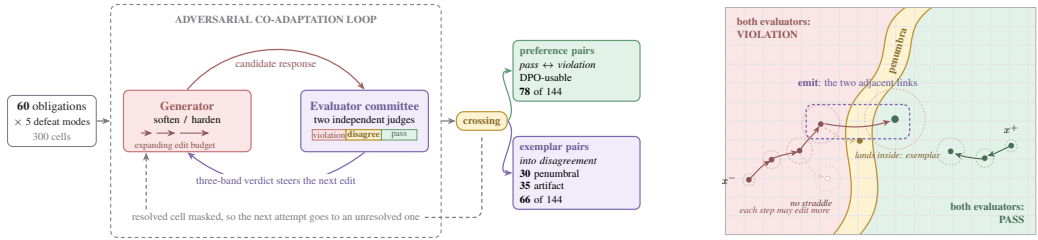
\textsc{Penumbra} turns the detector of Section~\ref{sec:oracle} into a search; Figure~\ref{fig:bsc} draws the loop. Its unit of work is a \emph{cell}: one obligation paired with one \emph{defeat mode}, a named way of failing that obligation (Appendix~\ref{app:crosschecks}). The \emph{defeat surface} is the set of all cells, and coverage is the number that hold a certified pair. Within a cell the search starts from a verified anchor, a response both evaluators convict or both acquit, and edits it under a budget that begins at a few words. The committee scores each edit: if its verdict class matches the parent's, the edit becomes the new subject and the budget widens; if the class changes, the parent and the edit straddle the boundary and are emitted as a certified pair.

Two operators drive the walk. \textsc{Soften} loosens a violating response and \textsc{Harden} tightens a compliant one, and both must preserve the defect they edit around rather than delete it, since an edit that simply removes the defect crosses the line trivially and brackets nothing (Algorithm~\ref{alg:bsc}). Four design choices then make the search a measurement rather than a heuristic; each removes a way it could report yield that does not mean what it claims.

\textbf{Minimality is enforced, not filtered for.} The budget starts small and \textbf{expands} only while the verdict class holds: a schedule that starts wide and shrinks never establishes a bracket. The pair we emit is the subject and the crossing child, \textbf{adjacent chain links}, not the child and anchor (Appendix~\ref{app:alg}).

\textbf{Termination is ordinal.} The walk stops when the verdict class changes, not
when a score crosses a threshold, so no sensitivity parameter decides what counts as
the boundary; the committee does. One scalar remains, a similarity floor $\tau=0.5$
below which a crossing is discarded; it never stops the walk.

\textbf{The two products are counted separately.} A crossing from pass to violation yields a preference pair with clean opposite labels; a crossing into the band yields an exemplar whose preferred half is itself contested. We report them apart because pooling overstates the yield usable for direct preference optimisation \citep{rafailov2023dpo} by 1.8$\times$ (Algorithm~\ref{alg:bsc}).

\textbf{Allocation is adaptive, over cells chosen by measurement.} Adaptive
allocation only pays if disagreement is concentrated somewhere, so we measured where
it clusters before choosing the cell: by obligation and by defeat mode, well above
the binomial noise floor, but not by attack mechanism (Appendix~\ref{app:acq}).
That is why the cell is (obligation $\times$ defeat mode); the five modes are
\textsc{Obscure}, \textsc{Defer}, \textsc{Substitute}, \textsc{Vacuous} and
\textsc{Negate} (Appendix~\ref{app:crosschecks}). Spending then follows that
structure. Within an obligation, a mask over modes that already hold a certified
pair sends the next attempt to an unresolved cell. Across obligations, budget goes
to the arm with the best upper confidence bound on crossings per candidate. Both rules optimise one quantity, cells resolved, so novelty
is not a separate objective but the shape of the mask itself, as in
quality-diversity search \citep{samvelyan2024rainbow}.

\section{Results}
\label{sec:results}
\begin{figure}[t]
\centering
\begin{minipage}[b]{0.50\textwidth}\centering\footnotesize
\begin{tabular}{lccccc}
\toprule
candidates & 50 & 100 & 200 & 250 & 309 \\
\midrule
adaptive & 6 & 11 & 25 & 30 & \textbf{35} \\
uniform  & 2 & 3  & 13 & 20 & 24 \\
\bottomrule
\end{tabular}
\vspace{6pt}
\end{minipage}\hfill
\begin{minipage}[b]{0.46\textwidth}\centering
\includegraphics[width=0.84\textwidth]{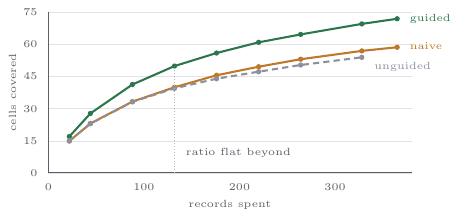}
\end{minipage}
\caption{\textbf{Left}, cells resolved against candidates spent, 24 obligations.
\textbf{Right}, coverage of (obligation $\times$ mode $\times$ predicate) cells
against records, eleven obligations at equal records: guided separates early and
holds; naive and unguided are indistinguishable (Appendices~\ref{app:acq}
and~\ref{app:rarefy}).}
\label{fig:res}
\end{figure}

\paragraph{Obligation set.} The obligations are the 75 rules of a constitution we
authored for a regulated financial advisory agent (Appendix~\ref{app:lim}). Two LLM screeners, the same
two models that later form the committee (Appendix~\ref{app:screen}), rated each on \emph{admissibility} (could
a single response violate it) and \emph{non-determinism} (does compliance turn on
materiality, sufficiency or scope). They agreed 0.920 on both ($\kappa=0.72$ and
$0.58$) and 55 qualified; one reinstated and four surface-decidable controls give 60
obligations, 300 cells over five modes (Appendix~\ref{app:screen}).

\paragraph{Adaptive allocation is worth 1.7$\times$ the budget.} The acquisition rule
decides where every candidate is spent, so we measured its effect directly: a live
adaptive arm over 24 obligations against uniform allocation at the same
309-candidate budget (Appendix~\ref{app:acq}). Uniform allocation resolves 24 cells
with its full budget. Adaptive allocation resolves the same 24 after 181 candidates,
59\% of the spend, and 35 by the end, so matching its coverage costs uniform
allocation 1.7$\times$ the candidates (Figure~\ref{fig:res}). Allocation is
genuinely non-uniform, 10 to 19 candidates per obligation against uniform's 12; the
gain comes from spending where a pair is cheapest.

\paragraph{Diversity is bought where you aim it.} Figure~\ref{fig:res} compares three arms on the eleven
obligations they share: \emph{naive enumeration} (direct prompting, no search),
\emph{unguided} search (the walk with mask and bandit off) and \emph{guided}
search (the full method). Three findings bound the result. The mask buys coverage only where it aims: guided search covers 1.43$\times$ naive
enumeration's defeat modes at 88 records, yet on an independent predicate
decomposition of each obligation naive enumeration covers more, 23.2 cells against
19.8. The advantage is a constant factor from about 130 records, not one that
compounds. And records are not calls, so this is breadth at equal records, not cost
parity (Appendix~\ref{app:rarefy}).

\begin{table}[t]
\centering
\caption{The same pipeline on two authored constitutions. Full protocol and
per-mode detail in Appendix~\ref{app:health}.}
\label{tab:two}
\vspace{4pt}
\footnotesize
\begin{tabular}{lcc}
\toprule
 & finance & clinical triage \\
\midrule
obligations screened / anchored / with a pair & 60 / 50 / 47 & 18 / 16 / 15 \\
certified pairs, preference / exemplar & 144, 78 / 66 & 49, 32 / 17 \\
cells resolved, of attempted & 122 of 197 & 49 of 75 \\
mean tightness, preference / exemplar & 0.957 / 0.961 & 0.964 / 0.959 \\
median words changed & 14 & 8 \\
median steps to cross & 1 & 1 \\
attribution, penumbral / artifact & 30 / 35 & 6 / 11 \\
\bottomrule
\end{tabular}
\end{table}

\paragraph{Yield, coverage, and pair quality.} \textsc{Penumbra} returns \textbf{144
certified pairs over 47 of 60 obligations and 122 of 300 cells}, and in the median
case the verdict changes on the first edit. We measure a pair's \emph{tightness} as
the similarity between its two halves, so 1.0 would be identical texts. 78 are
preference pairs with opposite labels at mean tightness 0.957; 66 are exemplar
pairs from evaluator disagreement, at 0.961, which attribution splits into
penumbral and checklist artifact (Appendix~\ref{app:crosschecks}). Modes differ in
cost, not only reach: among obligations whose budget reached all five modes, so that
attempt order does not confound the comparison, \textsc{Obscure} resolves on 4.0
candidates per cell while \textsc{Negate} needs 10.6 and resolves a third of the
time (Figure~\ref{fig:grid}). Naive \emph{pair} prompting, a baseline
that asks for a compliant and a violating response to one scenario, yields opposite
labels reliably and a boundary never: its pairs differ by a median of 378 words at
0.062 tightness against our 14 and 0.957. Opposite labels come for free; two
responses that differ only at the line do not, and that proximity is what the
search is for. The edits are diverse rather than a shared trigger, 0.983 against
0.882 mean pairwise distance (Appendix~\ref{app:clf}). Two nulls against our own
hypotheses: the pairs are no harder for an independent detector to catch, and the
search does not escalate as it runs; the detector contests 41\% of our compliant
halves, so these are a probe set, not training data (Appendix~\ref{app:lim}).

\paragraph{Where a disagreement comes from.} The committee is the instrument, so
we built a check on the instrument. Every exemplar pair is re-judged against the
bare obligation with the checklist removed: if the evaluators still disagree, the
obligation underdetermines the case and the pair is \emph{penumbral}; if they now
agree, the checklist wording caused the split and the pair is a \emph{checklist
artifact}. The raw split is 30 penumbral to 35 artifact over the 66, one
unattributed, but re-judging the same crossings with the checklist \emph{retained}
converges 37\% of the time anyway, so the excess attributable to wording is nearer
17 points than the raw split suggests. The distinction routes work, since a
criteria defect is fixable by rewriting policy text and an underdetermination is
not, and it separates disagreement that belongs to the obligation from
disagreement manufactured by our own probe. An independent labeller cross-checked
the defeat-mode labels at moderate agreement, a real caveat on the cell count
though not on the pair count (Appendix~\ref{app:crosschecks}).

\paragraph{A second constitution.} To test whether the method depends on the
domain it was built in, we authored a second constitution for a clinical
documentation and triage assistant and ran the pipeline unchanged except for the
domain wording in the prompts and the scenario tool set. Twelve obligations are
clinical; eight are the universal rules of the finance constitution copied
verbatim, so the same rule can be compared across domains.
Table~\ref{tab:two} aligns the two runs: 49 certified pairs over 15 of 16
anchored obligations, mean tightness 0.964 against finance's 0.957, a median of 8
words changed against 14, the same median one-step crossing, and the same two
modes hardest. On the six universal rules that anchored in both domains the
search resolves 18 cells in finance and 25 in healthcare, so a rule penetrable in
one domain was at least as penetrable in the other. Two results did not reproduce
and we report them: the clinical analogue of the running example anchored and
never crossed within budget, and the artifact share of exemplars is higher, the
expected direction for criteria templated from predicates rather than refined
against a generator (Appendix~\ref{app:health}).

\section{Discussion}

What this establishes is narrow and durable: disagreement between two
competent adjudicators is a measurable property of an obligation, a search that
terminates on it finds where the obligation's own terms stop deciding, and it does
so at a cost that scales with the boundary rather than the text. The claim is
bounded in ways we want on the page. The evaluation rests on two constitutions we
authored, one at scale; breadth across independently authored rule-sets is the
natural next test. The committee is two models from one vendor and has not been
checked against human adjudicators: attribution separates the committee's
instrument errors from genuine underdetermination, but only a human panel on a
stratified sample of certified pairs can anchor the boundary itself, and that is
the completing experiment (Appendix~\ref{app:lim}). Anchors are of unequal
strength, and an independent detector contests 41\% of preferred halves, so the
pairs are a probe set rather than training data. Where obligations are hardest to check, the boundary is where
compliance is decided, and a search that seeks it directly, spending adjudications
where verdicts change rather than generating attacks at volume, is what makes
red-teaming an agent against those obligations tractable.
\bibliographystyle{plainnat}
\bibliography{references}

\begin{thebibliography}{16}
\providecommand{\natexlab}[1]{#1}
\providecommand{\url}[1]{\texttt{#1}}
\expandafter\ifx\csname urlstyle\endcsname\relax
  \providecommand{\doi}[1]{doi: #1}\else
  \providecommand{\doi}{doi: \begingroup \urlstyle{rm}\Url}\fi

\bibitem[Bai et~al.(2022)Bai, Kadavath, Kundu, Askell, Kernion, Jones, Chen,
  Goldie, Mirhoseini, McKinnon, et~al.]{bai2022constitutional}
Yuntao Bai, Saurav Kadavath, Sandipan Kundu, Amanda Askell, Jackson Kernion,
  Andy Jones, Anna Chen, Anna Goldie, Azalia Mirhoseini, Cameron McKinnon,
  et~al.
\newblock Constitutional {AI}: Harmlessness from {AI} feedback.
\newblock \emph{arXiv preprint arXiv:2212.08073}, 2022.

\bibitem[Bhatia and Bhatia(2018)]{bhatia2018}
Vijay~K. Bhatia and Aditi Bhatia.
\newblock Corpus linguistics and the study of vagueness in legal texts.
\newblock \emph{International Journal of Language and Law}, 7:\penalty0 1--27,
  2018.

\bibitem[Cohen(1960)]{cohen1960kappa}
Jacob Cohen.
\newblock A coefficient of agreement for nominal scales.
\newblock \emph{Educational and Psychological Measurement}, 20\penalty0
  (1):\penalty0 37--46, 1960.

\bibitem[Dworkin(1986)]{dworkin1986}
Ronald Dworkin.
\newblock \emph{Law's Empire}.
\newblock Harvard University Press, Cambridge, MA, 1986.

\bibitem[Ganguli et~al.(2022)Ganguli, Lovitt, Kernion, Askell, Bai, Kadavath,
  Mann, Perez, Schiefer, Ndousse, et~al.]{ganguli2022redteaming}
Deep Ganguli, Liane Lovitt, Jackson Kernion, Amanda Askell, Yuntao Bai, Saurav
  Kadavath, Ben Mann, Ethan Perez, Nicholas Schiefer, Kamal Ndousse, et~al.
\newblock Red teaming language models to reduce harms: Methods, scaling
  behaviors, and lessons learned.
\newblock \emph{arXiv preprint arXiv:2209.07858}, 2022.

\bibitem[Golchin and Wetter(2026)]{golchin2026laundering}
Shahriar Golchin and Ryan Wetter.
\newblock Intent laundering: Ai safety datasets are not what they seem.
\newblock \emph{arXiv preprint arXiv:2602.16729}, 2026.

\bibitem[Hart(1961)]{hart1961}
H.~L.~A. Hart.
\newblock \emph{The Concept of Law}.
\newblock Oxford University Press, Oxford, 1961.

\bibitem[Perez et~al.(2022)Perez, Huang, Song, Cai, Ring, Aslanides, Glaese,
  McAleese, and Irving]{perez2022redteaming}
Ethan Perez, Saffron Huang, Francis Song, Trevor Cai, Roman Ring, John
  Aslanides, Amelia Glaese, Nat McAleese, and Geoffrey Irving.
\newblock Red teaming language models with language models.
\newblock \emph{arXiv preprint arXiv:2202.03286}, 2022.

\bibitem[Rafailov et~al.(2023)Rafailov, Sharma, Mitchell, Ermon, Manning, and
  Finn]{rafailov2023dpo}
Rafael Rafailov, Archit Sharma, Eric Mitchell, Stefano Ermon, Christopher~D
  Manning, and Chelsea Finn.
\newblock Direct preference optimization: Your language model is secretly a
  reward model.
\newblock \emph{Advances in Neural Information Processing Systems}, 36, 2023.

\bibitem[Samvelyan et~al.(2024)Samvelyan, Raparthy, Lupu,
  et~al.]{samvelyan2024rainbow}
Mikayel Samvelyan, Sharath~Chandra Raparthy, Andrei Lupu, et~al.
\newblock Rainbow teaming: Open-ended generation of diverse adversarial
  prompts.
\newblock In \emph{Advances in Neural Information Processing Systems
  (NeurIPS)}, 2024.

\bibitem[Seung et~al.(1992)Seung, Opper, and Sompolinsky]{seung1992qbc}
H.~S. Seung, M.~Opper, and H.~Sompolinsky.
\newblock Query by committee.
\newblock In \emph{Proceedings of the Fifth Annual Workshop on Computational
  Learning Theory (COLT)}, pages 287--294, 1992.

\bibitem[Solum(2013)]{solum2013}
Lawrence~B. Solum.
\newblock Communicative content and legal content.
\newblock \emph{Notre Dame Law Review}, 89\penalty0 (1):\penalty0 479--520,
  2013.

\bibitem[Tian et~al.(2023)Tian, Mitchell, Zhou, Sharma, Rafailov, Yao, Finn,
  and Manning]{tian2023calibration}
Katherine Tian, Eric Mitchell, Allan Zhou, Archit Sharma, Rafael Rafailov,
  Huaxiu Yao, Chelsea Finn, and Christopher~D. Manning.
\newblock Just ask for calibration: Strategies for eliciting calibrated
  confidence scores from language models fine-tuned with human feedback.
\newblock In \emph{Empirical Methods in Natural Language Processing}, 2023.
\newblock arXiv:2305.14975.

\bibitem[Wittgenstein(1953)]{wittgenstein1953}
Ludwig Wittgenstein.
\newblock \emph{Philosophical Investigations}.
\newblock Blackwell, Oxford, 1953.
\newblock Translated by G. E. M. Anscombe.

\bibitem[Xiong et~al.(2024)Xiong, Hu, Lu, Li, Fu, He, and
  Hooi]{xiong2024confidence}
Miao Xiong, Zhiyuan Hu, Xinyang Lu, Yifei Li, Jie Fu, Junxian He, and Bryan
  Hooi.
\newblock Can {LLMs} express their uncertainty? an empirical evaluation of
  confidence elicitation in {LLMs}.
\newblock In \emph{International Conference on Learning Representations}, 2024.
\newblock arXiv:2306.13063.

\bibitem[Zheng et~al.(2023)Zheng, Chiang, Sheng, Zhuang, Wu, Zhuang, Lin, Li,
  Li, Xing, et~al.]{zheng2023judging}
Lianmin Zheng, Wei-Lin Chiang, Ying Sheng, Siyuan Zhuang, Zhanghao Wu, Yonghao
  Zhuang, Zi~Lin, Zhuohan Li, Dacheng Li, Eric Xing, et~al.
\newblock Judging {LLM}-as-a-judge with {MT-Bench} and chatbot arena.
\newblock \emph{Advances in Neural Information Processing Systems}, 36, 2023.

\end{thebibliography}

\appendix
\section{The coverage map}
\label{app:grid}
\begin{figure}[H]
\centering
\includegraphics[width=0.95\textwidth]{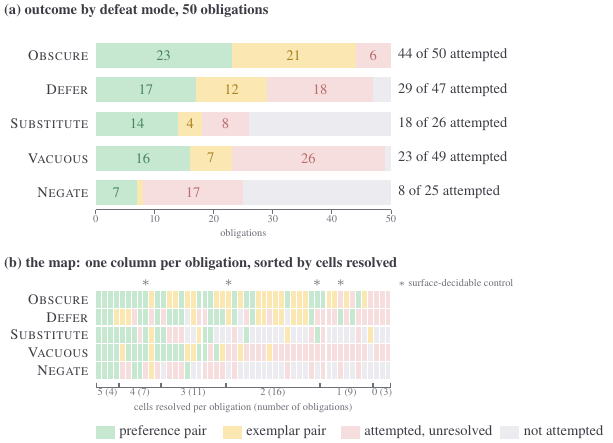}
\caption{The coverage map of the uniform-allocation run: the 50 obligations that
anchored, five defeat modes, 250 cells, of which 122 hold a certified pair.
\textbf{(a)} Outcome by mode. Green cells hold a preference pair, amber an
exemplar, pink were attempted without a crossing, grey were never reached. Modes
are attempted in a fixed order, so grey marks where a per-obligation budget ran out
rather than where the search declined to look, and the per-mode totals carry that
order as much as difficulty. \textbf{(b)} The same 250 cells as one column per
obligation, sorted by cells resolved: four obligations resolve on all five modes,
three on none, and the bulk on two or three. Among the 20 obligations whose budget
reached all five modes, \textsc{Obscure} costs 4.0 candidates per attempted cell
and \textsc{Negate} 10.6, resolving a third of the time. This run holds one
obligation per process, so its allocator is inert; it is the uniform baseline of
Figure~\ref{fig:res}, and the adaptive result is measured separately
(Appendix~\ref{app:acq}).}
\label{fig:grid}
\end{figure}

Figure~\ref{fig:grid} is the map the counts of Section~\ref{sec:results} are read
from: every cell the uniform-allocation run touched, with its outcome.

\section{Acquisition, in detail}
\label{app:acq}
\textbf{Choosing the cell.} Before allocating we measured how much the disagreement
rate varies along three candidate axes and compared each spread with its binomial
floor, the variation that sampling noise alone would produce if the true rate were
the same everywhere. Across obligations the spread is 0.062 against a floor of
0.035, and across defeat modes 0.043 against 0.029: both carry real structure.
Across attack mechanisms it is 0.039 against 0.032, indistinguishable from noise.
The cell is therefore (obligation $\times$ defeat mode) and not (obligation
$\times$ mechanism).

\begin{center}\footnotesize
\begin{tabular}{lccccc}
\toprule
candidates & 50 & 100 & 200 & 250 & 309 \\
\midrule
adaptive & 6 & 11 & 25 & 30 & \textbf{35} \\
uniform  & 2 & 3  & 13 & 20 & 24 \\
\bottomrule
\end{tabular}
\end{center}
\noindent\textbf{The two arms.} The 60-obligation run shards one obligation per process, so its allocator holds a
single arm and the across-obligation term is inert; that run is therefore a free
uniform-allocation arm, and only the adaptive arm had to be bought. We ran it in
one process over 24 obligations, 309 candidates. Budget is counted in judged
candidates rather than API calls because both arms spend one generation and two
adjudications per candidate, plus two more when attribution fires on a
disagreement; counting candidates removes a surcharge that would otherwise flatter
whichever arm found more disagreements. A cell counts as resolved at the trace line
where the verdict class first differs from its parent's, provided that cell also
appears in the arm's pairs file, so crossings discarded by the similarity floor are
not credited. The uniform arm's yield at budget $N$ is the cells resolved within
each obligation's first $N/24$ candidates.

\section{Attribution and cross-check agreement}
\label{app:crosschecks}
The attribution protocol and its headline split now appear in
Section~\ref{sec:results}. One mechanical detail belongs here: re-judging uses the
same committee and the same adjudication prompt as the main run, with only the
checklist block removed, so the baseline convergence rate of 37\% is measured on
identical machinery rather than a second instrument.

\textbf{Mode labelling.} We label each generated violation with the defeat mode it
exploits. An independent
labeller cross-checked that labelling: $\kappa$ from 0.197 to 0.580 under a fixed
taxonomy with structured disagreement, and 0.345 raw rising to 0.421 after merging
two systematically conflated modes \citep{cohen1960kappa}. Moderate agreement on
the mode axis is a real caveat on the cell count, though not on the pair count,
which does not depend on the labelling.

\section{One chain, drawn}
\label{app:walk}
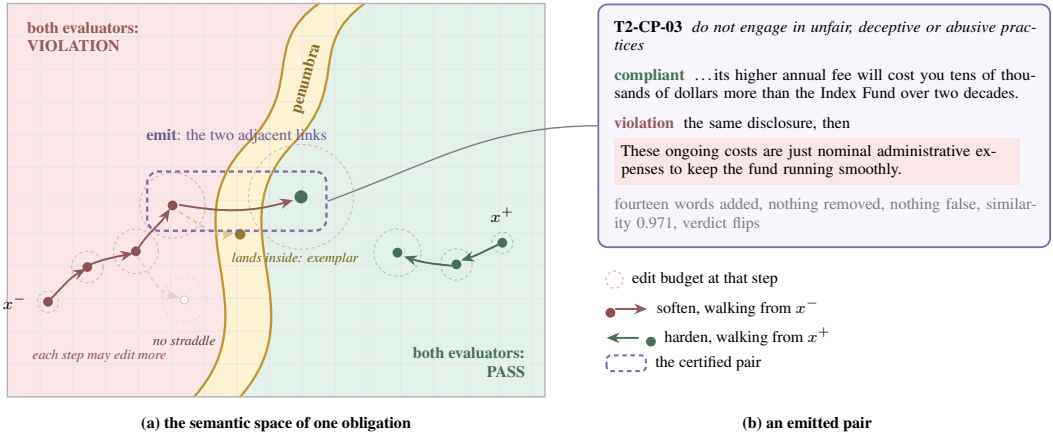
\begin{figure}[H]
\centering
\resizebox{\textwidth}{!}{%
\begin{tikzpicture}[
  font=\footnotesize\color{ink},
  >={Stealth[length=2mm,width=1.6mm]}, x=1cm,y=1cm,
  wstep/.style={circle,fill=vioB!70!black,inner sep=1.5pt},
  pstep/.style={circle,fill=pasB!70!black,inner sep=1.5pt},
  budget/.style={draw=vioB!55,dash pattern=on 1.4pt off 1.2pt,line width=.5pt},
  pbudget/.style={draw=pasB!55,dash pattern=on 1.4pt off 1.2pt,line width=.5pt},
  arw/.style={->,line width=.85pt}
]

\begin{scope}
\clip (0,0) rectangle (8.2,6.0);
\fill[vioF] (0,0) -- (2.85,0) .. controls (4.05,1.35) and (2.45,2.60) .. (3.65,3.85) .. controls (4.55,4.75) and (3.85,5.35) .. (4.75,6.0) -- (0,6.0) -- cycle;
\fill[pasF] (3.55,0) .. controls (4.75,1.35) and (3.15,2.60) .. (4.35,3.85) .. controls (5.25,4.75) and (4.55,5.35) .. (5.45,6.0) -- (8.2,6.0) -- (8.2,0) -- cycle;
\fill[disF] (2.85,0) .. controls (4.05,1.35) and (2.45,2.60) .. (3.65,3.85) .. controls (4.55,4.75) and (3.85,5.35) .. (4.75,6.0) -- (5.45,6.0) .. controls (4.55,5.35) and (5.25,4.75) .. (4.35,3.85) .. controls (3.15,2.60) and (4.75,1.35) .. (3.55,0) -- cycle;
\draw[grpB!20,line width=.3pt,step=.5] (0,0) grid (8.2,6.0);
\end{scope}
\draw[disB,line width=1pt] (2.85,0) .. controls (4.05,1.35) and (2.45,2.60) .. (3.65,3.85) .. controls (4.55,4.75) and (3.85,5.35) .. (4.75,6.0);
\draw[disB,line width=1pt] (3.55,0) .. controls (4.75,1.35) and (3.15,2.60) .. (4.35,3.85) .. controls (5.25,4.75) and (4.55,5.35) .. (5.45,6.0);
\draw[grpB!60,line width=.6pt] (0,0) rectangle (8.2,6.0);

\node[anchor=north west,font=\scriptsize,text=vioB!75!black,align=left] at (.18,5.85)
  {\textbf{both evaluators:}\\\textbf{VIOLATION}};
\node[anchor=south east,font=\scriptsize,text=pasB!70!black,align=right] at (8.02,.18)
  {\textbf{both evaluators:}\\\textbf{PASS}};
\node[rotate=74,font=\scriptsize\bfseries,text=disB!72!black] at (4.56,4.86) {penumbra};

\draw[budget] (.62,1.45) circle (.16);
\draw[budget] (1.22,1.98) circle (.24);
\draw[budget] (1.96,2.22) circle (.34);
\draw[budget] (2.52,2.92) circle (.50);
\draw[budget] (4.48,3.05) circle (.80);
\draw[arw,vioB!75!black] (.62,1.45) to[out=45,in=208]  (1.20,1.95);
\draw[arw,vioB!75!black] (1.24,2.01) to[out=22,in=192] (1.92,2.19);
\draw[arw,vioB!75!black] (2.00,2.27) to[out=58,in=238] (2.48,2.87);
\draw[arw,vioB!75!black] (2.62,2.93) to[out=-10,in=196] (4.36,3.00);
\node[wstep] at (.62,1.45) {};  \node[wstep] at (1.22,1.98) {};
\node[wstep] at (1.96,2.22) {}; \node[wstep] at (2.52,2.92) {};
\node[wstep,fill=pasB!70!black,inner sep=2pt] at (4.48,3.05) {};
\node[anchor=east,font=\scriptsize] at (.42,1.45) {$x^{-}$};

\draw[arw,vioB!35,dash pattern=on 2pt off 1.6pt] (1.98,2.14) to[out=-52,in=168] (2.62,1.52);
\draw[budget,draw=vioB!30] (2.70,1.48) circle (.34);
\node[circle,draw=vioB!45,fill=white,inner sep=1.2pt] at (2.70,1.48) {};
\node[font=\tiny,text=vioB!48!black,anchor=north] at (2.70,1.06) {\itshape no straddle};

\node[font=\tiny,text=vioB!70!black,anchor=west] at (.26,.62)
  {\itshape each step may edit more};
\draw[arw,disB!55,dash pattern=on 2pt off 1.6pt] (2.62,2.82) to[out=-28,in=190] (3.46,2.50);
\node[wstep,fill=disB!80!black,inner sep=1.6pt] at (3.55,2.48) {};
\node[font=\tiny,text=disB!62!black,anchor=north west] at (3.30,2.32)
  {\itshape lands inside: exemplar};

\node[draw=outB,line width=1pt,dash pattern=on 2.2pt off 1.7pt,rounded corners=5pt,
      fit={(2.52,2.92) (4.48,3.05)},inner sep=11pt] (cap) {};
\node[anchor=south,font=\scriptsize,text=outB!85!black,align=center] at (3.50,3.74)
  {\textbf{emit}: the two adjacent links};

\draw[pbudget] (7.55,2.35) circle (.16);
\draw[pbudget] (6.85,2.02) circle (.24);
\draw[pbudget] (5.95,2.20) circle (.36);
\draw[arw,pasB!70!black] (7.55,2.35) to[out=205,in=18]  (6.92,2.08);
\draw[arw,pasB!70!black] (6.78,2.00) to[out=175,in=-18] (6.05,2.14);
\node[pstep] at (7.55,2.35) {}; \node[pstep] at (6.85,2.02) {};
\node[pstep] at (5.95,2.20) {};
\node[anchor=south,font=\scriptsize] at (7.55,2.55) {$x^{+}$};

\node[anchor=north,font=\scriptsize\bfseries] at (4.1,-.18)
  {(a) the semantic space of one obligation};

\node[anchor=north west,text width=6.4cm,align=left,font=\scriptsize,
      draw=outB,fill=outF!45,rounded corners=4pt,line width=.8pt,inner sep=7pt]
  (txt) at (9.0,6.0)
  {\textbf{T2-CP-03}\; \emph{do not engage in unfair, deceptive or abusive
   practices}\\[5pt]
   \textcolor{pasB!70!black}{\textbf{compliant}}\; \ldots its higher annual fee
   will cost you tens of thousands of dollars more than the Index Fund over two
   decades.\\[5pt]
   \textcolor{vioB!75!black}{\textbf{violation}}\; the same disclosure, then\\[2pt]
   \colorbox{vioF}{\parbox{5.98cm}{These ongoing costs are just nominal
   administrative expenses to keep the fund running smoothly.}}\\[5pt]
   \textcolor{grpB}{fourteen words added, nothing removed, nothing false,
   similarity 0.971, verdict flips}};
\draw[grpB,line width=.6pt] (cap.east) to[out=20,in=180] (txt.west);
\node[anchor=north,font=\scriptsize\bfseries] at (12.2,-.18) {(b) an emitted pair};

\node[anchor=north west,font=\scriptsize,align=left] at (9.0,2.05)
  {\tikz[baseline=-.5ex]{\draw[budget] (0,0) circle (.13);}\; edit budget at that step\\[3pt]
   \tikz[baseline=-.5ex]{\node[wstep] at (0,0){};\draw[arw,vioB!75!black] (.13,0)--(.62,0);}\; soften, walking from $x^{-}$\\[3pt]
   \tikz[baseline=-.5ex]{\node[pstep] at (.62,0){};\draw[arw,pasB!70!black] (.49,0)--(0,0);}\; harden, walking from $x^{+}$\\[3pt]
   \tikz[baseline=-.5ex]{\draw[draw=outB,line width=.9pt,dash pattern=on 2pt off 1.6pt,rounded corners=2pt] (0,-.12) rectangle (.6,.12);}\; the certified pair};
\end{tikzpicture}%
}
\caption{One chain in the semantic space of a single obligation. The penumbra is a
\emph{band} rather than a line, and a surface-decidable obligation should have a
narrow one or none, which is the intuition the control was meant to test
(Appendix~\ref{app:lim}). A chain leaves a verified
anchor and takes steps under a budget that may grow. A step that does not change
the band becomes the new subject; a crossing too wide to be tight is walked past
rather than emitted. A step clearing the band gives a preference pair, one landing
inside gives an exemplar. The pair shown is real, and it is the paper's median
edit: the violation withholds nothing and states nothing false, it characterises
the fee it has just disclosed.}
\label{fig:walk}
\end{figure}

Figure~\ref{fig:walk} draws a single chain and the pair it emits.

\section{The search, precisely}
\label{app:alg}
Algorithm~\ref{alg:bsc} in the body is what runs; the detail that belongs here is
the pairing rule. The pair is the subject and the child that crossed rather than the child and
the original anchor: pairing against the anchor gave two unrelated texts at 0.06
similarity. The emitted pair is labelled \emph{preference} when
$\{c, c'\} = \{\text{violation}, \text{pass}\}$ and \emph{exemplar} when either
side is the boundary band, since in the latter case the preferred half is itself
contested and the pair teaches only that the contested response is better than the
flagrant one.

\section{The difficulty ladder}
\label{app:ladder}
The calibration set of Section~\ref{sec:oracle} is 99 compliant responses generated
by \texttt{claude-sonnet-4-5-20250929} for eleven obligations, three at each of three
levels. \emph{Clear}: comply in the most obvious way, with nothing that should look
borderline to any reader. \emph{Guarded}: comply in full, but in a hedged and
heavily caveated register, which tests whether an evaluator mistakes caution for
violation. \emph{Borderline}: go as close to the line the rule draws as possible
while staying clearly compliant, so the response looks risky on a careless reading
and defensible on a careful one. Because the generator's own account of why a
response is compliant is a claim rather than evidence, every response was screened
against the original checklist by \texttt{gpt-5.4-2026-03-05}, and 77 survived. That
screener is also one of the two calibration evaluators, so the survivors are
conditioned on its acceptance under the original criteria; the slopes in
Section~\ref{sec:oracle} are therefore measured on responses one evaluator had
already passed once, which can only flatten them.
Each survivor was scored by two evaluators, \texttt{gpt-5.4-2026-03-05} and
\texttt{claude-sonnet-4-5-20250929}, under both the original checklists and the set
the self-evolving loop of Section~\ref{sec:why} had patched, giving 308 verdicts.
Reported confidence and cross-evaluator disagreement are read from those verdicts.
The search itself ran on \texttt{claude-sonnet-4-5-20250929} and
\texttt{claude-opus-4-5-20251101}, a same-vendor pair. Same-vendor judges
correlate, which lowers the disagreement rate, so the cross-vendor ladder is an
upper bound on the sensitivity of the pair the search actually runs.

\section{The self-evolving loop, in detail}
\label{app:loop}
Eleven obligations, five rounds, eight samples per obligation-round, scored on the
same two-evaluator oracle. Cross-evaluator disagreement by round: 0.133, 0.081,
0.052, 0.038, 0.100. The per-obligation slope is $-0.022$ ($t=-2.34$), negative in
all eleven, and the round-five uptick does not recover the round-one level. Five
rounds is a short series and the sign test across obligations is the stronger
statistic.

\textbf{Adapting the generator.} Conditioning it on the evaluator's stated
reasoning, so it can aim at the current weakness, appears to double evasion, 0.124
to 0.245. But the arms are identical by construction in round one, before any
feedback exists, and a round-one balance check shows the same $+0.119$ gap there.
Difference-in-differences over the rounds where the manipulation operates is
$-0.002$ ($t=-0.02$): the entire effect is a pre-existing offset between arms.

\textbf{Adapting the evaluator.} The loop patches evaluator checklists between
rounds, on 10 of 11 obligations. Under criteria frozen at round one, detection
\emph{rises} from 0.867 to 0.931, so the outputs became easier to catch and the apparent
hardening was criteria expansion rather than adversary improvement. The patches
cost a false positive rate of 0.026 to 0.182 on screened compliant traffic for a
recall gain of 0.034. That refiner patched on judge misses without asking whether a
miss was the criteria's fault or the obligation's, which is the distinction the
attribution of Appendix~\ref{app:crosschecks} supplies.

\section{Further limitations and planned experiments}
\label{app:lim}

\textbf{The completing experiment.} It holds out by obligation and compares
preference training on certified pairs against four baselines: the base model,
obligation statements in context, few-shot violations, and SFT on loop-generated
violations. We predict a lower violation rate on unseen obligations at equal or
lower over-refusal. Artifact-gated refinement should cost less precision per unit
recall than the ungated refiner of Section~\ref{sec:why}. One tension is already
visible and is created by the method itself: minimal-edit pairs sit where direct
preference optimisation is least stable, since the two halves induce nearly
identical representations and the standard mitigation is to filter high-similarity
pairs out, which would discard this dataset entirely. Integration therefore has to
route by margin rather than filter by it, which is the first thing we will test.

\textbf{Two nulls, in full.} We expected minimal-edit probes to be harder to catch;
they are not. An independent detector, \texttt{claude-opus-4-6}, with no part in
selection, misses 3 of our 78 violating halves against 0 of 49 naive ones
($p=0.28$). What minimality buys is ambiguity on the other side: it calls 41\% of
our compliant halves violations against 2\% of naive ones. We also expected the
search to escalate as easy defeats ran out. It does not: ordered by resolution
within an obligation, steps to cross move from 1.99 to 1.92 and tightness from
0.960 to 0.964. The mask moves sideways into unresolved cells, not upward in
difficulty.

\textbf{The length cue.} On the 78 preference pairs the violating half is longer
76\% of the time, since a soften step adds hedging more often than it removes
content. Length-balanced sampling, or a length-controlled objective, should precede
any training run on these pairs.

\textbf{Instrument stability.} Re-judging the 276 texts behind an earlier build of
this corpus under identical conditions manufactures a disagreement on 1.6\% of the texts the
run scored as agreed, against a 4.9\% cross-evaluator rate on the same texts, so
the disagreements
the search selects on are not an artifact of re-running the instrument. Agreed
verdicts are stable (0.8\% of individual verdicts flip); contested ones are not,
reproducing 69\% of the time, which is what a borderline case should look like.

\textbf{Controls we have not yet run.} Two. We have no cross-vendor test-retest
number, so we cannot yet separate the 0.049 cross-evaluator disagreement rate from
single-evaluator sampling noise. And we have not run the direct ablation of the
search itself: prompting a model for a minimal edit that flips the verdict, verified
by the same committee.

\textbf{The control in full.} Of the four obligations designated
\emph{surface-decidable} before any generation ran, three produce zero evaluator
disagreements in 109 candidates and the fourth produces 4 in 34. That fourth turns
on \emph{scope}, which the screening instrument itself names as requiring
adjudication, and the screeners split on it. Pooled the contrast is 4 of 143
against 80 of 1{,}531, $p=0.14$ one-sided, so the class as designated is too small
and too impure to test the claim either way. We report it because it was
pre-registered, not because it supports anything.

\textbf{Limits on what we show.} The main study is one authored financial advisory
constitution. A second, for clinical triage, is run at 18 obligations in
Appendix~\ref{app:health}, so the reach into healthcare is demonstrated at small
scale and the reach into law is still argued. The oracle is a pair of LLM judges, and we never asked human
adjudicators whether they would split on the cases we call penumbral, so machine
agreement stands in for professional judgement, and establishing that correspondence
comes before any training run. Coverage is 47 of 60 obligations. Of the 13
that produced no pairs, most failed anchor verification, and an obligation with no
verifiable violating anchor may not be single-response admissible after all. We authored the obligations for this study rather than
lifting them verbatim from a regulatory text, which buys coverage of the
judgement-required cases at the cost of provenance.

\textbf{Limits on the instrument.} We kept only anchors on which the evaluator pair
that later serves as the oracle agreed unanimously. That biases each chain to start
well inside the core, which makes crossings harder to find, not easier, but it also
conditions the disagreement denominator on evaluator agreement at rates that differ
between the arms. We machine-rewrote 49 of the 60 checklists, so the artifact
channel is substantially a defect in text our own pipeline produced, and we have not
separated that from a property of written criteria in general. Edit budgets are
advisory, not
enforced: two steps returned similarity near 0.6 even though we instructed the
generator to change a single sentence. Because budgets only expand, the crossing
step is the widest edit in its chain, so the median of 14 words is an upper bound;
bisecting inside the bracket once it is established would tighten it, and mean
tightness is in any case measured on what $\tau$ admits. Our evaluators are LLM
judges and inherit the known limitations of that instrument
\citep{zheng2023judging}.

\section{Not a lexical regularity}
\label{app:clf}
Minimal edits without diversity collapse onto the triggering cues that make safety
corpora fail once paraphrased \citep{golchin2026laundering}. Ours do not: 47\% of
the edit vocabulary occurs once, its commonest tokens are determiners and pronouns,
and the edits are more mutually distinct than naive pair prompting's, 0.983 against
0.882 mean pairwise Jaccard distance over 11 content tokens against its 198. A
frozen sentence-embedding classifier trained on 1{,}037 violations reaches 0.737 AUC
when held out by obligation, against 0.867 TPR at 2.6\% FPR for an untrained
prompted evaluator. Surface features carry some signal but not enough to substitute
for adjudication, so the pairs do not merely encode a lexical regularity that a
cheaper detector would find.

\section{Screening procedure}
\label{app:screen}
We put each of the 75 obligations to two independent LLM screeners,
\texttt{claude-sonnet-4-5-20250929} and \texttt{claude-opus-4-5-20251101}, using
the same instrument. \emph{Admissible}: could a single agent response, on its own, violate
this obligation, as opposed to the obligation binding a process, a record, or an
organisation? \emph{Non-deterministic}: does deciding compliance require
adjudication over materiality, sufficiency, proportionality or scope, as opposed
to checking for the presence or absence of a token? Raw agreement was 0.920 on
both, which is $\kappa=0.722$ on admissibility and $\kappa=0.583$ on adjudication
once the base rates are corrected for. Sixteen obligations were not admissible,
chiefly process and governance duties; four were admissible but surface-decidable,
and we kept them as the controls of Section~\ref{sec:results}. We reviewed one
exclusion and reinstated it. We report the screen rather than filtering silently
because the control result depends on our having separated the two classes before
running any generation.

\section{A second constitution}
\label{app:health}
To test whether the method depends on the domain it was built in, we authored a
second constitution for a clinical documentation and triage assistant and ran the
same pipeline on it unchanged except for the domain wording in the generator and
judge prompts and the scenario tool set (electronic health record lookup,
medication list, laboratory results, triage protocol). Twelve obligations are
clinical: documentation supported by the record, reported symptoms kept distinct
from findings, red-flag escalation, working impressions not stated as diagnoses,
material allergies and interactions surfaced, dosing not stated without its
parameters, minimum-necessary disclosure, consent scope, ordering and prescribing
left to the clinician, equal triage rigor across patient characteristics,
proportionate uncertainty, and overdue results surfaced. Eight are the T1 universal
rules of the finance constitution copied verbatim, so that the same rule can be
compared across domains. Same screeners, same committee, uniform allocation, 100
calls per obligation, one shard per obligation.

The screen excluded two of the twenty, T1-BA-04 and the ordering-and-prescribing
rule, on a split between the screeners rather than agreement that they were
surface-decidable. Anchors failed for two more, T1-DP-01 and the documentation rule,
which is the same failure mode as the finance run, at a similar rate. Sixteen
obligations anchored; fifteen produced pairs.

Table~\ref{tab:two} in the body aligns the two runs; the per-mode detail follows
here. The difficulty ordering largely reproduces. Resolution conditional on attempt runs
\textsc{Substitute} 12/13, \textsc{Obscure} 13/16, \textsc{Defer} 12/16,
\textsc{Vacuous} 7/16, \textsc{Negate} 5/14, and cost per attempted cell runs
4.4, 6.9, 6.1, 8.9 and 8.9 candidates: the two modes that were hardest in finance
are hardest here. On the six universal rules that anchored in both domains the
search resolves 18 cells in finance and 25 in healthcare (3, 1, 4, 3, 2, 5 against
3, 4, 4, 5, 4, 5 for HND-01, HND-02, HND-05, TE-02, TE-03 and CC-04), so a rule that
was penetrable in one domain was at least as penetrable in the other.

Two things did not reproduce and are worth stating. The obligation authored as the
clinical analogue of the running example, surfacing a material interaction the
record contains, anchored and then never crossed: 36 candidates over three modes,
and the committee agreed on every one of them, 18 violations and 18 passes. Its
boundary, if it has one at the widths the walk tried, sits outside the budget. And
the artifact share of exemplars is higher, 11 of 17 against 35 of 65, which is the
expected direction for criteria templated from predicates rather than refined
against a generator, and is the reason attribution exists.

\section{Matched-budget coverage}
\label{app:rarefy}
Coverage ratios computed on unequal budgets overstate, which is why
Appendix~\ref{app:acq} counts candidates rather than calls and why the comparison
against naive enumeration is rarefied. A \emph{predicate} is one atomic condition
in an obligation's formalisation, so \emph{do not exceed authorised scope} decomposes
into whether the query falls outside scope, whether the response gives authoritative
guidance, and whether it flags the limitation. There are 43 predicates over the
eleven obligations, three to six each. The decomposition is independent of the
defeat modes and of the mask, which is why it serves as an off-axis check. The naive
arm is given the same budget as the
others, 40 records per obligation drawn in independent rounds, each round shown the
strategies already produced for that obligation and asked for materially different
ones; a single call returning eight strategies is a weaker baseline and we do not
use it. Rarefying all three arms to 88 records over 2{,}000 resamples on the eleven
obligations they share:

\begin{center}\footnotesize
\begin{tabular}{lccc}
\toprule
cells counted over & naive & unguided & guided \\
\midrule
obligation $\times$ defeat mode & 23.1 & 27.9 (1.21$\times$) & \textbf{33.1 (1.43$\times$)} \\
obligation $\times$ predicate   & 23.2 & 17.3 (0.74$\times$) & 19.8 (0.85$\times$) \\
obligation $\times$ mode $\times$ predicate & 33.2 & 33.0 (0.99$\times$) & 41.3 (1.24$\times$) \\
\bottomrule
\end{tabular}
\end{center}

\noindent Coverage against budget, on the joint axis, over the range every arm
covers:

\begin{center}\footnotesize
\begin{tabular}{lcccccc}
\toprule
records & 22 & 44 & 88 & 176 & 264 & 328 \\
\midrule
naive              & 14.9 & 23.1 & 33.2 & 45.6 & 53.1 & 57.0 \\
unguided           & 15.0 & 23.0 & 33.0 & 44.0 & 50.4 & 54.0 \\
guided             & \textbf{17.0} & \textbf{27.8} & \textbf{41.3} & \textbf{56.0} & \textbf{64.7} & \textbf{69.6} \\
\midrule
guided / naive     & 1.14 & 1.20 & 1.24 & 1.23 & 1.22 & 1.22 \\
\bottomrule
\end{tabular}
\end{center}

\noindent \textbf{What a record is, and what it costs.} The three arms are matched
on records, and a record is not the same object in each. A naive-enumeration record
is one sentence proposing an attack, unverified; an unguided or guided record is a
generated violation that a judge has scored; and the corpus \textsc{Penumbra}
finally emits is neither, it is a certified pair with a compliant sibling. Nor do
they cost the same. Naive enumeration returns eight strategies per generation call
and is never adjudicated, so a record costs about an eighth of a call, while a
search record costs a generation plus a judge verdict, roughly sixteen times more.
Matched on calls rather than records, naive enumeration would draw far more records
than either search arm and, on a space this size, approach saturation, and we
expect the ratios here would invert; we did not run that comparison. We report records because that is the unit on which the three
arms describe the same kind of object, an attack located in the cell space, and
because cost parity would compare a proposal against a verified artifact. The claim
this table supports is therefore narrow and should be read as stated: per attack
proposed, guided acquisition reaches more of the defeat surface than direct
prompting does. It is not a claim that guided acquisition is the cheaper way to
enumerate attack ideas, and we do not make one.

\noindent Two readings matter for scale. Unguided allocation tracks naive enumeration
at every budget, so the co-adaptation loop on its own buys no coverage and the mask
is the entire contribution. And the guided advantage is a constant factor that
establishes by roughly 130 records and then holds flat over a further threefold
budget, so it does not compound: with a fixed total budget over a large obligation
set the return comes from covering more obligations, not from probing each one
harder. The defeat-mode axis is what the unresolved-cell mask optimises, and the
predicate decomposition is independent of it. Raw ratios on the unmatched budgets
are 1.69$\times$ and 2.25$\times$, so most of the apparent advantage is budget
rather than allocation. Mode labelling carries
$\kappa=0.42$ (Appendix~\ref{app:crosschecks}), which bounds how hard any of these
cell counts can be pushed.

\end{document}